\documentclass{article}

\usepackage{geometry}
\usepackage[english]{babel}
\usepackage[utf8]{inputenc}
\usepackage{johd}

\usepackage{geometry} 

\usepackage{setspace}

\def\ie{\emph{i.e.}}

\usepackage{natbib}
\usepackage{makecell}
\usepackage{amsmath,amssymb,amsfonts}
\usepackage{graphicx}
\usepackage[table,xcdraw]{xcolor}
\usepackage{array}
\usepackage{booktabs}
\usepackage{multirow}
\usepackage{caption}
\usepackage{subcaption}
\usepackage{xcolor}
\usepackage{soul}
\usepackage{lineno}

\definecolor{electricindigo}{rgb}{0.44, 0.0, 1.0}
\definecolor{lightblue}{RGB}{240,245,255}
\definecolor{darkblue}{RGB}{40,40,85}
\definecolor{babyblue}{rgb}{0.54, 0.81, 0.94}
\definecolor{pearDark}{HTML}{2980B9}
\definecolor{pearDarker}{HTML}{1D2DEC}
\definecolor{columbiablue}{rgb}{0.61, 0.87, 1.0}

\newcommand{\tabincell}[2]{\begin{tabular}{@{}#1@{}}#2\end{tabular}}  

\title{MammoDG: Generalisable Deep Learning Breaks the Limits of Cross-Domain Multi-Center Breast Cancer Screening}

\author{Yijun Yang$^{a,b}$$^{*}$, Shujun Wang$^{a,e}$, Lihao Liu$^{a}$, Sarah Hickman$^{c,d}$, Fiona J Gilbert$^{d}$,\\ Carola-Bibiane Schönlieb$^{a}$, Angelica I. Aviles-Rivero$^{a}$\\ \\
        \small $^{a}$DAMTP, University of Cambridge, Cambridge, UK \\
        \small $^{b}$ROAS, The Hong Kong University of Science and Technology (Guangzhou), Guangzhou, China \\
        \small $^{c}$Department of Radiology, Barts Health NHS Trust, The Royal London Hospital, UK \\
        \small $^{d}$Department of Radiology, Biomedical Research Centre,  University of Cambridge, Cambridge, UK \\
        \small $^{e}$The Department of Biomedical Engineering, The Hong Kong Polytechnic University, Hong Kong, China \\\\
        \small $^{*}$Corresponding author: Yijun YANG; \tt{yyang018@connect.hkust-gz.edu.cn}\\  {\small Work done while interning at the University of Cambridge.} \\
}
\date{}

\begin{document}

\maketitle

\begin{abstract} 
Breast cancer is a major cause of cancer death among women, emphasising the importance of early detection for improved treatment outcomes and quality of life. Mammography, the primary diagnostic imaging test, poses challenges due to the high variability and patterns in mammograms. Double reading of mammograms is recommended in many screening programs to improve diagnostic accuracy but increases radiologists' workload. Researchers explore Machine Learning models to support expert decision-making.  Stand-alone models have shown comparable or superior performance to radiologists, but some studies note decreased sensitivity with multiple datasets, indicating the need for high generalisation and robustness models. This work devises MammoDG, a novel deep-learning framework for generalisable and reliable analysis of cross-domain multi-center mammography data. MammoDG leverages multi-view mammograms and a novel contrastive mechanism to enhance generalisation capabilities.  Extensive validation demonstrates MammoDG's superiority, highlighting the critical importance of domain generalisation for trustworthy mammography analysis in imaging protocol variations.
\end{abstract}

\section{Introduction}\label{sec1_intro}

Breast cancer is the second leading cause of cancer death in women worldwide\footnote{\url{https://www.cancer.org/cancer/types/breast-cancer/about/how-common-is-breast-cancer.html}}. Early cancer detection is relevant for treatment and improvement of life quality and outcomes. Mammography is the primary imaging test for diagnosis yet its interpretation is a major challenge~\citep{marmot2013benefits,pharoah2013cost}.  The number of false-positive and false-negative findings is due to the high variability and patterns in the mammograms.  Therefore, it is often necessary to advocate a double reading of mammograms, which increases radiologists’ workload, cost, and time~\citep{royal2019clinical}.

Some prior research has been devoted to developing Machine Learning (ML) models to support expert decision-making and achieve comparable to or superior performance to radiologists with stand-alone  tools~\citep{rodriguez2019stand,mckinney2020international}. 
However, in other studies, the sensitivity performance is observed to decrease or without change when facing large cohorts from different sites and dataset characteristics~\citep{schaffter2020evaluation}. 
The reason is that the large cohort contains the out-of-distribution (OOD) data collected from different vendor machines and protocols in different sites leading to a distribution shift of the imaging data.

The body of literature on ML for mammography cancer diagnosis can be broadly divided into three main categories: 1) single view-based models~\citep{zhu2017deep,wu2019deep,yala2019deep}; 2) multiple view-based models~\citep{geras2017high,khan2019multi,zhao2020cross,wei2022beyond}; and 3) patch-based techniques~\citep{mercan2017multi,agarwal2019automatic,wu2019deep}. 
Moreover, these models can use either a single ML model or ensembles.
However, they do not include any mechanism or are designed to address the above problem of distribution shift from large cohorts of mammography data. Whilst the ML community has studied this topic with domain generalization for other real-world applications~\citep{zhou2022domain}, the works on domain generalisation for analysing mammograms are scarce. In recent work, \cite{li2021domain} uses contrastive learning principles to further augment the generalization capability of the deep learning model considering 4 seen vendors and one unseen vendor. However, that approach still is limited in terms of extracting more richer and statistical information.

In this work, we address the challenging question of  -- how to design deep learning models that can be generalisable, robust, and reliable  to multi-center OOD data.
With this purpose in mind, we introduce a novel deep learning framework based on domain generalisation to mitigate the distribution shift problem, on mammography screening tasks, namely MammoDG. Our new framework considers multi-view mammograms. The key of our framework is how we harmonise richer statistical information from multiple views and enforce fine-grained detection  via a proposed contrastive mechanism. Our contributions are summarised as follows. 

\begin{figure}[h]%
\centering
\includegraphics[width=\textwidth]{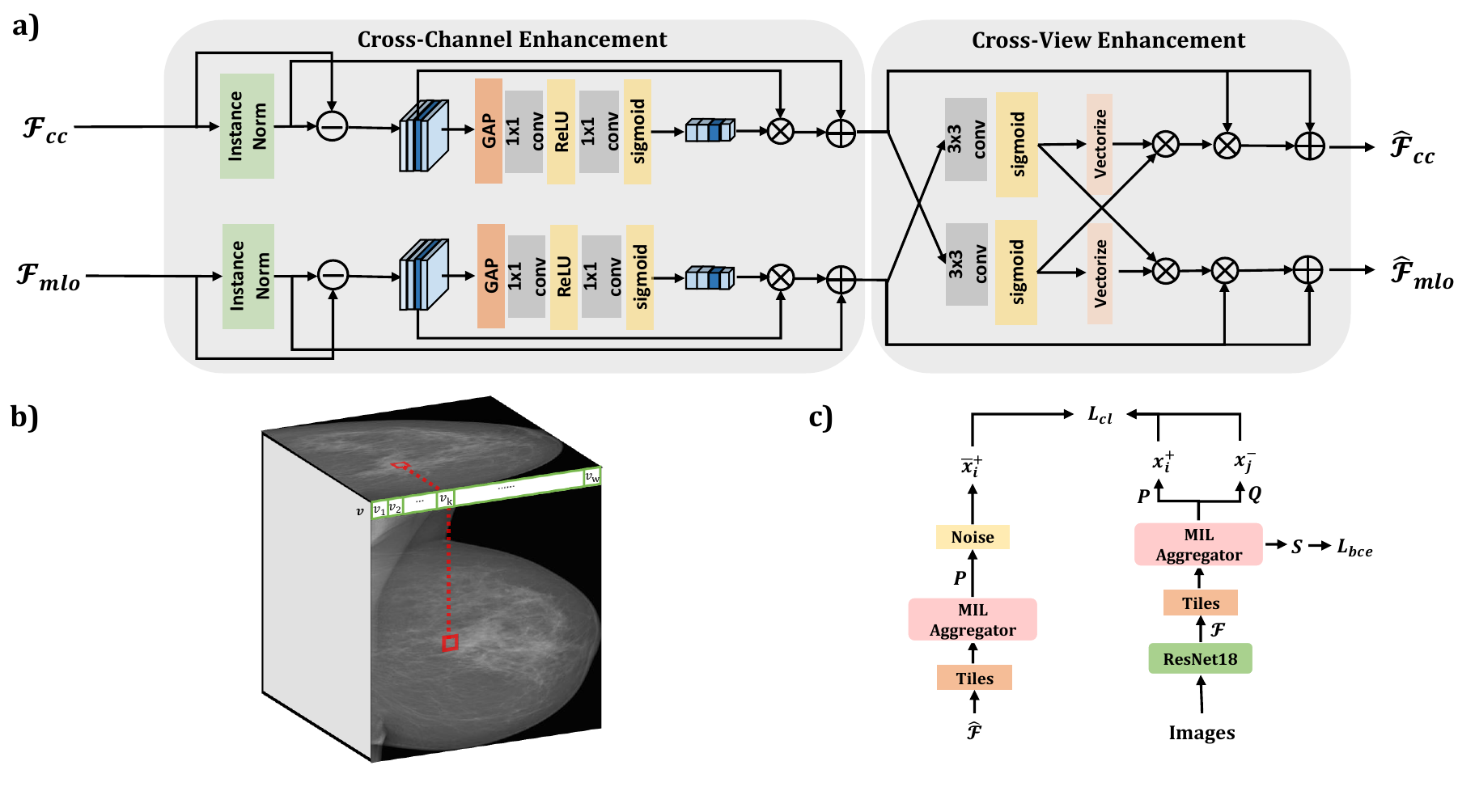}
\caption{\textbf{(a) CVE module}. First, task-relevant features are distilled from the input feature $\mathcal{F}$ to achieve Cross-Channel Enhancement for each view. Secondly, in Cross-View Enhancement, the geometric-attended vector $\textbf{v}_{mlo}$ computed from the channel-enhanced feature $\tilde{\mathcal{F}}^+_{mlo}$ is multiplied by the self-attention map of the CC view to integrate the complementary information from the MLO view into the feature of CC view. \textbf{(b) The visualisation of our geometric-attended vector $\textbf{v}$} helps understand the principle of Cross-View Enhancement. The attended value of abnormal tissues in the $k$-th column in CC view is summarised in $v_k$ to provide valuable geometric information for the corresponding column in MLO view. \textbf{(c) Multi-Instance Contrastive Learning strategy.} $\hat{\mathcal{F}}$ is a mini-batch of enhanced feature maps by CVE module obtained from ResNet18 while $\mathcal{F}$ is the same mini-batch of original feature maps from ResNet18.}
\label{fig:cve}
\end{figure}

\begin{enumerate}
    \item[(i)] We propose a novel domain generalisation framework, MammoDG (Figure \ref{fig:framework}), for
    breast-level mammography diagnosis (classification). We highlight an interpretable multi-view strategy with a Cross-Channel Cross-View Enhancement module (Figure \ref{fig:cve}(a)). This module seeks to effectively harmonise the statistical information 
    from CC and MLO views in the middle feature phase (Figure~\ref{fig:cve}(b)).
    
    \item[(ii)] We introduce 
    a novel 
    Multi-instance Contrastive Learning mechanism (MICL) to enhance generalisation and fine-grained detection capabilities of our model. Our mechanism enforces local and global knowledge to address the out-of-distribution samples drawn from different vendors and hospitals large-scale acquisitions (as shown in Figure~\ref{fig:cve}(c)).  
    
    \item[(iii)] We extensively validate our new framework using benchmarking and in-home datasets from different vendor machines and sites, three of which are seen and two of which are unseen. We demonstrate that our model leads to better performance than existing deep learning models by a large margin on both seen and unseen datasets.
     \item[(iv)] We have shown that domain generalisation is critical to ensure trustworthiness and reliable deep learning models for mammography analysis, where limited data and substantial variations across imaging protocols and vendors machines.
\end{enumerate}
\section{Methods}\label{sec2_methods}

\begin{figure}[t!]%
\centering
\includegraphics[width=\textwidth]{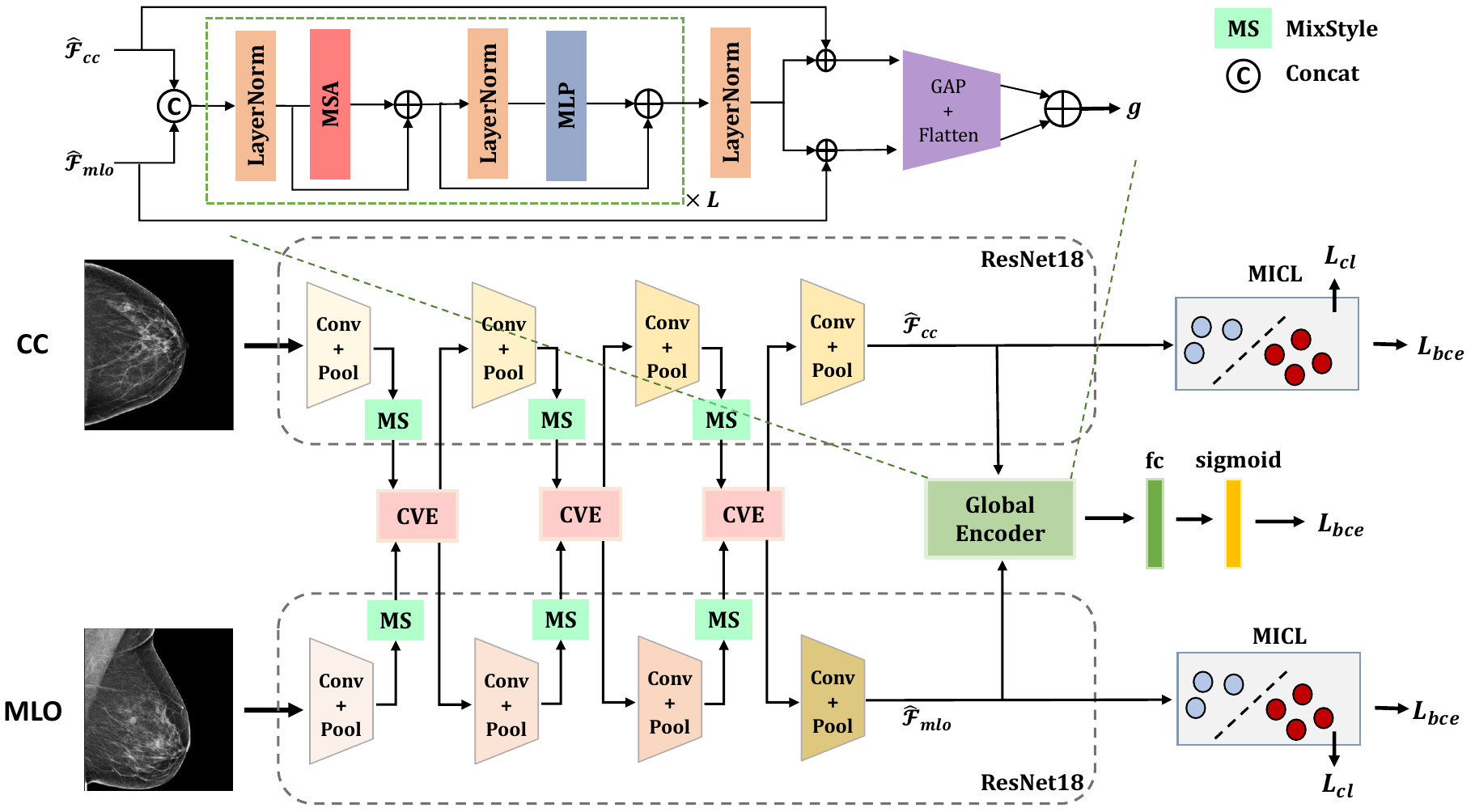}
\caption{
\textbf{{Overview} of our 
MammoDG {framework}.}
{A batch of CC and MLO pair views, from different domains, are fed into two-stream view-specific learning networks.} Our CVE modules learn their statistical knowledge from the same pair, at the first three levels, while global encoder further integrates two-stream feature maps $\hat{\mathcal{F}_{cc}},\hat{\mathcal{F}_{mlo}}$ at the last level. The share decoder, consisting of two fully connected layers and a sigmoid layer, generates breast-level predictions. To give a strong supervision by discovering patch information across domains, MICL plays view-specific learning and generates image-level predictions.}
\label{fig:framework}
\end{figure}

In this section, we describe in detail our proposed  MammoDG framework for addressing the out-of-distribution problem in breast cancer screening.
Figure~\ref{fig:framework} depicts our domain generalisation framework for breast-level mammography classification.
%
We consider a training set of multiple source domains $\mathcal{S} = \{S_1,...,S_K\}$,  where each domain $S_k$ contains $N_k$ weakly labelled samples ${(a_i^k, b_i^k, y_i^k)}_{i=1}^{N_k}$ representing the CC and MLO views, and breast-level labels respectively.
%
Our framework learns a domain-agnostic model, $f_\theta: X\rightarrow Y$, using $K$ distributed source domains so that it can 
generalise to a completely unseen domain $\mathcal{T}$ 
without performance degradation. 

The CC and MLO views are first fed into two-stream view-specific learning networks to obtain their multi-level feature representations. 
A Cross-\textbf{C}hannel Cross-\textbf{V}iew \textbf{E}nhancement (\textbf{CVE}) module is then proposed to learn the data statistical 
knowledge. We also introduce a Transformer as global encoder for better final feature fusion.
%
The view-specific and shared decoder subnetworks are then adopted to provide image-level and breast-level predictions. \textit{To extract domain-invariant features from data from different vendors}, we propose \textbf{M}ulti-\textbf{I}nstance \textbf{C}ontrastive \textbf{L}earning (\textbf{MICL}), which uses the principles of Multi-Instance Learning and Contrastive Learning for boosting performance by detecting abnormal critical instances (patches) across domains.

\subsection{Cross-Channel Cross-View Enhancement}\label{subsec41}
Previous work in Multi-view Mammography Classification either adopted a single-stream network to separately process different views~\citep{shen2021interpretable,li2021domain}, or directly concatenate the outputs of the multi-stream network in the late fusion level~\citep{geras2017high,wu2019deep,khan2019multi}. However, existing works do not consider the statistical information shared by two views, of the same breast, 
at the middle feature level. To this end, we introduce a CVE module to enhance the feature representation of one view by exploiting complementary knowledge from the other view. The CVE includes two parts, \ie, cross-channel and cross-view feature enhancement, as illustrated in Figure~\ref{fig:cve}(a).
First, we leverage Instance Normalisation (IN) to perform style normalisation by normalising feature statistics from different distributions (domains). While IN has the power to better the generalisation ability of networks, it inevitably results in weaker discrimination capability. To recover task-relevant discriminative feature from the IN removed information, we conduct cross-channel enhancement. Specifically, we distill the task-relevant feature from the residual feature $\mathcal{R}$ of the original feature $\mathcal{F}$ and the normalised feature $\tilde{\mathcal{F}}$, which reads: $\mathcal{R} = \mathcal{F}-\tilde{\mathcal{F}}.$
We highlight task-relevant part $\mathcal{R}^+$ from $\mathcal{R}$ through a learned channel-wise attention vector $\textbf{t}=[t_1,t_2,...,t_C]\in \mathbb{R}^C$:
\begin{equation}
    \begin{aligned}
    &\mathcal{R}^+(:,k) = t_k\mathcal{R}(:,k),\\
    \textbf{t} &= \sigma(\theta_2\delta(\theta_1\text{GAP}(\mathcal{R}))),\\
    \end{aligned}
\label{eq:channelattention}
\end{equation}
where the attention module is implemented by a spatial global average pooling layer (GAP), followed by two $1\times 1$ convolutional layers (that are parameterised by $\theta_1\in \mathbb{R}^{c\times(c/r)}$ and $\theta_2\in \mathbb{R}^{(c/r)\times c}$), $\sigma(\cdot)$ and $\delta(\cdot)$ represent sigmoid activation function and ReLU  function, respectively. To reduce the number of parameters, a dimension reduction ratio $r$ is empirically set to 16. After that, we obtain the channel-enhanced feature by adding the distilled task-relevant feature $\mathcal{R}^+$ to the normalised feature $\tilde{\mathcal{F}}$ as:
\begin{equation}
    \begin{aligned}
    \tilde{\mathcal{F}}^+ &= \tilde{\mathcal{F}} + \mathcal{R}^+.
    \end{aligned}
\label{eq:residual}
\end{equation}

Once we have obtained the channel-enhanced feature representations from different views, one critical task is to effectively integrate them. Intuitively, as the CC and MLO views capture the same breast from above and side, abnormal tissues in the same breast can be observed in both views. To exploit the correlations between the two views, we propose using the geometric-attended vector. Specifically, we calculate their feature-level attention maps by a $3\times 3$ convolutional layer ($\theta_3$) with a sigmoid function as
\begin{equation}
    \begin{aligned}
    w_{cc} = \sigma(\theta_3(\tilde{\mathcal{F}}^+_{cc})),\ 
     w_{mlo} = \sigma(\theta_3(\tilde{\mathcal{F}}^+_{mlo})),\\
    \end{aligned}
\label{eq:channelattention}
\end{equation}
We then aggregate the complementary information into a learned column-wise geometric-attended vector $\textbf{v}=[v_1,v_2,...,v_W]\in \mathbb{R}^W$ to enhance the other view. We regard the maximum weight of each column,  $w$, as the summarised value in our geometric-attended vector $\textbf{v}$. For example, as Figure~\ref{fig:cve}(b) shows, the abnormal tissue in the $k$-th column of CC view should exist in the corresponding column of the MLO view, and thus the geometric information is summarised in $v_k$ by assigning the bigger attended value. After obtaining the geometric-attended vector, we multiply it by the attention map of the other view to differentiate the pixels in the same column. This process reads:
\begin{equation}
    \begin{aligned}
    \hat{w}_{cc} = w_{cc}\cdot \textbf{v}_{mlo},
     \hat{w}_{mlo} = w_{mlo}\cdot \textbf{v}_{cc}.\\
    \end{aligned}
\label{eq:channelattention}
\end{equation}

Finally, we achieve cross-view enhancement by adding the attended feature to the input feature as
\begin{equation}
    \begin{aligned}
    \hat{\mathcal{F}}_{cc} = \tilde{\mathcal{F}}^+_{cc} + \hat{w}_{cc}\cdot \tilde{\mathcal{F}}^+_{cc},\ 
    \hat{\mathcal{F}}_{mlo} = \tilde{\mathcal{F}}^+_{mlo} + \hat{w}_{mlo}\cdot \tilde{\mathcal{F}}^+_{mlo}.\\
    \end{aligned}
\label{eq:residual}
\end{equation}
The cross-channel cross-view enhanced feature representation $\hat{\mathcal{F}}$ is propagated to the next layer of each stream network to capture and integrate multi-level information.

\subsection{Multi-Instance Contrastive Learning}\label{subsec42}
Regions of interest (ROI) in mammography images, such as masses, asymmetries, and microcalcifications, are often small and sparsely distributed over the breast, and may present as subtle changes in the breast tissue pattern. The Multiple Instance Learning (MIL) technique is a great solution to improve fine-grained detection when ROI annotations are not available~\citep{zhu2017deep}. However, due to the absence of global guidance, the instance classifier is much more likely to be confused by local knowledge in patches from images of different distributions. It is hard to fully leverage MIL when samples come from different domains. On the other hand, \cite{li2021domain} recently proposed employing self-supervised Contrastive Learning to attain the goal of generalization robustness in mammography detection tasks. However, they depend on CycleGAN~\citep{zhu2017unpaired} to generate multi-style multi-view images, which may have poor quality due to unexpected changes in tiny tissues of patches.


To address these limitations, we propose Multi-Instance Contrastive Learning (MICL) scheme by integrating MIL and Contrastive Learning to more effectively enhance both the generalization and fine-grained detection capability of the model. As Figure~\ref{fig:framework} shows, we treat our MICL module as view-specific decoder subnetworks to preserve the special knowledge of each view while the shared information can be learned in the shared decoder. The detailed procedure of MICL is illustrated in Figure~\ref{fig:cve}(c). Specifically, we adopt a dual-stream MIL aggregator~\citep{li2021dual} to jointly learn a patch (instance) and an image (bag) classifier. 
Before feeding the cross-channel cross-view enhanced feature map $\hat{\mathcal{F}}$ to MICL, we divide it into $n\times n$ tiles along the spatial dimension to generate the bag of $n^2$ instances. 
Let $B=\{p_1,...,p_{n^2}\}$ denotes a bag of instances of one view. 
The MIL aggregator first decides the critical instance $p_m$ in a bag by using the instance classifier $f_m(\cdot)$ on each instance embedding $p_i$ and max-pooling the scores. This process is given by:
\begin{equation}
    \begin{aligned}
    x = p_m &= \mathop{argmax}\limits_{p_i \in B}f_m(p_i),\\
    S_m(B) &= \mathop{max}\limits_{p_i \in B}f_m(p_i).
    \end{aligned}
\label{eq:maxpool}
\end{equation}
Secondly, the MIL aggregator measures the distance between each instance and the critical instance $p_m$, and then produces a bag embedding by summing the instance embeddings using the distances as weights. More specifically, each instance embedding $p_i$ (including critical instance $p_m$) is transformed into two vectors, query $q_i$ and information $v_i$, by linear layers. The distance $d_i$ denotes the similarity between queries of the instance embedding $p_i$ and critical instance embedding $p_m$, which is calculated by inner product and softmax. The bag score is further given by the bag classifier $f_b(\cdot)$:
\begin{equation}
    \begin{aligned}
    S_b&(B) = f_b(\sum_{i}^{n^2}d_iv_i).\\
    \end{aligned}
\label{eq:maxpool}
\end{equation}
The final score $S(B)$ is the average of the scores of the dual streams:
\begin{equation}
    \begin{aligned}
    S(B) &= \frac{1}{2}(S_m(B)+S_b(B)).\\
    \end{aligned}
\label{eq:maxpool}
\end{equation}

As the critical instance represents its bag and plays a significant role in both streams, it is necessary to guide the network to select the correct instance in a bag. To this end, we integrate weakly-supervised contrastive learning into multiple instance learning. First of all, we separate the critical instances of bags in a mini-batch into the malignant set $P=\{x^+_i\mid y_i=1\}$ and the benign set $Q=\{x^-_j\mid y_j=0\}$ according to breast-level labels. Then for each malignant critical instance $x^+_i$ as an anchor, we adopt its out-of-distribution view $\bar{x}^+_i$ as the positive sample while all benign critical instances are negative samples. 
Instead of standard data augmentation that cannot perturb the distribution of images and may destroy details in breast tissues, we apply a feature-level augmentation protocol comprised of Mixstyle~\citep{zhou2021domain} and random noise on the whole feature maps to obtain out-of-distribution instance embeddings. Mixstyle is inserted between layers in the CNN architecture to perturbing the distribution information of images from source domains inspired by Adaptive Instance Normalization. More specifically, given an input batch of feature maps $\textbf{F}$ and the shuffled batch $\textbf{F}^{'}$, Mixstyle computes their feature statistics, \ie, the mean $\gamma(\textbf{F}),\gamma(\textbf{F}^{'})$ and variance $\beta(\textbf{F}),\beta(\textbf{F}^{'})$. Then, we mix their feature statistics by linear interpolation:
\begin{equation}
    \begin{aligned}
    \gamma_{mix} = m\gamma(\textbf{F})+(1-m)\gamma(\textbf{F}^{'}),\ 
    \beta_{mix} = m\beta(\textbf{F})+(1-m)\beta(\textbf{F}^{'}),\\
    \end{aligned}
\label{eq:maxpool}
\end{equation}
where $m$ is randomly sampled from the uniform distribution, $m\sim U(0,1.0)$. Finally, the mixture of feature statistics is applied to the distribution-normalized $\textbf{F}$:
\begin{equation}
    \begin{aligned}
    \textbf{F}_{mix} = \beta_{mix}\cdot\frac{\textbf{F}-\gamma(\textbf{F})}{\beta(\textbf{F})}+\gamma_{mix}.\\
    \end{aligned}
\label{eq:maxpool}
\end{equation}
Note that we randomly shuffle the order in the batch dimension of $\textbf{F}$ to obtain $\textbf{F}^{'}$. Mixstyle only perturbs the distribution information of images, promising that the correlations among patches from one image remain invariant. 
Based on $\textbf{F}_{mix}$, we additionally inject slight feature noise to alleviate over-fitting.

Similar to InfoNCE contrastive loss~\citep{oord2018representation}, we applied our modified contrastive loss on the sampled features to give a stronger and more stable supervision:
\begin{equation}
    \begin{aligned}
    \mathcal{L}_{cl}=-\frac{1}{\left\vert P \right\vert}\sum_{x^+_i\in P}\log{\frac{e^{h(x^+_i)\cdot h(\bar{x}^+_i)/\tau}}{e^{h(x^+_i)\cdot h(\bar{x}^+_i)/\tau}+\sum_{x^-_j\in Q}e^{h(x^+_i)\cdot h(x^-_j)/\tau}}},\\
    \end{aligned}
\label{eq:clloss}
\end{equation}
where $\left\vert P \right\vert$ is the cardinality of $P$, $\tau$ is a scalar temperature hyper-parameter, and $h(\cdot)$ denote global average pooling and the normalization operation to convert instance embeddings into normalized feature vectors. Finally, the view-specific objective function of our MICL can be formulated as:
\begin{equation}
    \begin{aligned}
    \mathcal{L}_{cc}=\mathcal{L}_{bce}(S_{cc}(B^k_i),y^k_i) + \lambda \mathcal{L}_{cl},\ 
    \mathcal{L}_{mlo}=\mathcal{L}_{bce}(S_{mlo}(B^k_i),y^k_i) + \lambda \mathcal{L}_{cl},\\
    \end{aligned}
\label{eq:clloss}
\end{equation}
where $L_{bce}(\cdot)$ is binary cross entropy for supervised learning, and $\lambda$ is a balancing hyper-parameter.

Our MICL scheme has several inherent advantages compared with the original MIL and self-supervised Contrastive Learning:
(1) \textbf{Hard negative mining}: The selection of negative samples is crucial for learning contrastive features effectively~\citep{kalantidis2020hard}. Instead of including all instances in a bag into contrastive learning, we only consider the critical instance that has the highest score. This naturally provides our MICL with the ability to mine hard negative samples since the critical instance is most likely to be the false positive in a negative bag.
(2) \textbf{Mini-batch training}: To improve the generalisation robustness, we ensure that the mini-batch is composed of all source domains evenly during training. Our MICL can effectively suppress the confusion caused by patches from different domains not only because negative samples come from diverse distributions but also because Mixstyle enforces positive samples to contain the distribution information of negative samples, making the model more focus on task-related information.

\subsection{Global Encoder}\label{subsec43}

After MICL enforces view-specific learning, we aggregate the feature maps $\hat{\mathcal{F}}$ from CC and MLO branches using Transformer as the final global encoder to incorporate the global context for two views due to their complementary nature, as shown in Figure~\ref{fig:framework}. 
Specifically, we introduce Transformer~\citep{vaswani2017attention} to apply a multi-head self-attention mechanism, and operate on grid structured feature maps to discover the spatial dependencies between patches. 
Let the grid-structured feature map of a single view be a 3D tensor with dimensions $H\times W\times C$. 
For CC and MLO views, their features are stacked together to form a sequence with dimension $(2\times H\times W ) \times C$. We add a learnable positional embedding, which is a trainable parameter with dimension $(2\times H\times W ) \times C$, so allow the network to infer spatial dependencies between different tokens during training.
The input sequence and positional embedding are combined using element-wise summation to form a tensor of dimension $(2\times H\times W ) \times C$ as the input of the transformer. The output is then reshaped into two feature maps of dimension $H\times W\times C$ and fed back into each branch using an element-wise summation with the existing feature maps.

To save computational cost, we downsample higher resolution feature maps using average pooling to a fixed resolution of $H = W = 16$ before passing them as inputs to the transformer and upsample the output to the original resolution using bilinear interpolation before element-wise summation with the existing feature maps. After Transformer, the feature map is converted into a 512-dimensional feature vector by global average pooling. The feature vectors from both views are further combined via element-wise summation. 
This final 512-dimensional feature vector $\textbf{g}$ constitutes a compact representation that encodes the global context of two views. This is then fed to the shared decoder subnetwork which consists of two fully connected layers($\theta_4$) to obtain the breast-level prediction. The objective function of the shared decoder subnetwork is formulated as:
\begin{equation}
    \begin{aligned}
    \mathcal{L}_{sh}=\mathcal{L}_{bce}(\sigma(\theta_4(\textbf{g}^k_i)),y^k_i).\\
    \end{aligned}
\label{eq:clloss}
\end{equation}
Finally, we formulate a unified and end-to-end trainable framework. The overall loss function can be formulated as follows:
\begin{equation}
    \begin{aligned}
    \mathcal{L}_{total}=\mathcal{L}_{sh}+L_{cc}+\mathcal{L}_{mlo}.\\
    \end{aligned}
\label{eq:clloss}
\end{equation}

\subsection{Implementation Details}\label{subsec44}

Our proposed framework was trained on an NVIDIA A100 GPU and implemented on the Pytorch platform. The backbone of our framework was first pre-trained on BI-RADS labels following \citep{shen2021interpretable} and then finetuned on our seen domains. Our framework was empirically trained for 50 epochs in an end-to-end manner and the Adam optimizer was applied. The initial learning rate was set to $2 \times 10^{-5}$ and decayed by 10\% every 5 epochs. During training, we first resized and randomly crop the mammography images to $512 \times 512$, and then applied the image augmentation protocol including random horizontal flipping (p=0.5), random rotation ($-15^\circ$ to $15^\circ$), random translation (up to 10\% of image size), scaling by a random factor between 0.8 and 1.6, random shearing ($-25^\circ$ to $25^\circ$), and pixel-wise Gaussian noise ($\mu=0, \sigma=0.005$).
A batch of 12 cases evenly composed of three seen domains (\ie, CBIS, CMMD, TOMMY1) was fed into the network each time.

\section{Results}\label{sec3_results}
In this section, we present a comprehensive account of all the experiments we conducted to validate our proposed MammoDG framework.

\subsection{Data Description}
In this study, we use four datasets to evaluate the model generalisation performance, including three public datasets: CBIS, CMMD, and INBreast, and one private dataset TOMMY. 
Due to the large size of the TOMMY dataset, we split it into two non-overlapping parts TOMMY1 and TOMMY2 based on the patient level.
To assess the generalisation ability of the models, all the datasets utilized in this study are split into the Seen domain and the Unseen domain.
Seen domain means that the datasets contain both training and testing samples, \ie, their training samples are seen to the model, while Unseen domain means that the whole dataset is utilised for only testing.
Experimentally, we regard the CBIS, CMMD, and TOMMY1 as the Seen domain, TOMMY2 and INBreast as the Unseen domain for the final performance evaluation. For model selection strategy, we chose the checkpoint with the best performance on Unseen domain as the final model.
Data splits for Seen domain were created at the breast level, meaning that exams from a given breast were all in the same split.

\paragraph{CBIS-DDSM dataset}
CBIS-DDSM \citep{lee2017curated} is a public database of scanned film mammography studies containing cases categorized as normal, benign, and malignant with verified pathology information. 
It is a collection of mammograms from Massachusetts General Hospital, Wake Forest University School of Medicine, Sacred Heart Hospital, and Washington University of St Louis School of Medicine.
Mammography image data from CBIS-DDSM is an updated version of the DDSM providing easily accessible data.
We followed the official splits but discarded the cases that did not have both CC and MLO views, resulting in 572 benign, 475 malignant for training and 153 benign, 102 malignant for testing.
We did not use any data from DDSM for testing given that it is a scanned film dataset.

 \paragraph{CMMD dataset}
CMMD \citep{cui2021chinese} is a large public mammography database collected from patients from China, categorized as benign and malignant with verified pathology information.
Mammography image data were acquired on a GE Senographe DS mammography system.  
We split the breast-level cases with complete views into 80\%/20\% training/model selection splits, resulting in 423 benign, 1,021 malignant studies for training and 115 benign, 246 malignant studies for testing.

 \paragraph{INBreast dataset}
INBreast \citep{moreira2012inbreast} is a small public mammography database with a relatively balanced benign and malignant case. 
We split data from patient-level into breast-level and excluded cases with incomplete views, resulting in 125 benign and 46 malignant out of 171 studies.

 \paragraph{TOMMY dataset}
TOMMY \citep{gilbert2015tommy} is a rich and well-labeled dataset with over 7,000 patients (over 1,000 malignant) collected through six NHS Breast Screening Program (NHSBSP) centers throughout the UK and read by expert radiologists. To keep the number of breast-level cases consistent with other datasets, we just sampled a part of TOMMY for experiments.
TOMMY1 as Seen domain has 1,560 benign cases and 364 malignant cases for training, 406 benign and 76 malignant cases for testing.
TOMMY2 with 2,108 benign and 394 malignant cases was all treated as Unseen domain. The TOMMY1 and TOMMY2 datasets were obtained from Hologic vendor machines. 

\begin{table*}[!tbp]
  \centering
  \caption{Quanlitative results of our method compared to many state-of-the-art methods on our setting. All the models are trained on the training sets of seen domains and evaluated on the test set of seen domains and the whole set of unseen domains. The best performance is highlighted in \colorbox{pink}{red} colour while the second best results are in \colorbox{columbiablue}{blue} colour.}
    \resizebox{1.0\textwidth}{!}{%

\begin{tabular}{c|c|c|ccccccc|c} 
\hline
\hline
\multicolumn{2}{c|}{\textbf{Datasets}}                                   & \textbf{Metric} &\textbf{BIRADS} & \textbf{DMV-CNN} & \textbf{MVFF} & \textbf{GMIC}                     & \textbf{\thead{MSVCL\\(ResNet)}} & \textbf{\thead{MSVCL\\(FCOS)}}               & \textbf{Baseline}                  & \textbf{Ours}                      \\ 

\hline
\multirow{20}{*}{\rotatebox[origin=c]{90}{\textbf{Seen Domain}}}   & \multirow{4}{*}{\textbf{CBIS}}     & AUC & 0.6660          & 0.6654           & 0.7344        & 0.7666                            & 0.6874                & 0.7045                            & 0.7544                            & \cellcolor[HTML]{FFCCC9}{{0.7798}}   \\
                                  &                                 & TPR  & 0.6078          & 0.6078           & 0.6536        &\cellcolor[HTML]{FFCCC9} {{0.6993}}  & 0.6209                & 0.6013                            & 0.6732                            & \cellcolor[HTML]{C9E4FF}{{0.6932}}  \\
                                  &                                    & TNR & 0.6176          & 0.6176           & 0.6569        & \cellcolor[HTML]{FFCCC9}{{0.7034}}  & 0.6275                & 0.6176                            & 0.6765                            & \cellcolor[HTML]{C9E4FF}{{0.6863}}  \\
                                  &                                 & ACC  & 0.6118          & 0.6118           & 0.6549        & \cellcolor[HTML]{FFCCC9}{{0.7020}}  & 0.6235                & 0.6078                            & 0.6745                            & \cellcolor[HTML]{C9E4FF}{{0.6884}}  \\ 
\cline{2-11}
                                  & \multirow{4}{*}{\textbf{CMMD}}  & AUC & 0.6661          & 0.6818           & 0.7686        & \cellcolor[HTML]{C9E4FF}{{0.8157}} & 0.7878                & 0.8070                            & 0.8018                            & \cellcolor[HTML]{FFCCC9}{{0.8181}}   \\
                                  &                                 & TPR  & 0.6087          & 0.6435           & 0.6957        & \cellcolor[HTML]{C9E4FF}{{0.7304}} & 0.7130                & 0.7217                            & 0.7291                            & \cellcolor[HTML]{FFCCC9}{{0.7391}}   \\
                                  &                                 & TNR  & 0.6098          & 0.6382           & 0.6870        & \cellcolor[HTML]{C9E4FF}{{0.7398}} & 0.7195                & 0.7236                            & 0.7217                            & \cellcolor[HTML]{FFCCC9}{{0.7439}}   \\
                                  &                                 &  ACC  & 0.6094          & 0.6399           & 0.6898        & \cellcolor[HTML]{C9E4FF}{{0.7368}} & 0.7175                & 0.7230                            & 0.7241                            & \cellcolor[HTML]{FFCCC9}{{0.7424}}   \\ 
\cline{2-11}
                                  & \multirow{4}{*}{\textbf{TOMMY1}} & AUC  & 0.6624          & 0.6977           & 0.7178        & 0.7146                            & 0.7039                & \cellcolor[HTML]{FFCCC9}{{0.7535}}  & 0.6665                            & \cellcolor[HTML]{C9E4FF}{{0.7235}}  \\
                                  &                                 & TPR  & 0.5936          & 0.6108           & 0.6576        & 0.6404                            & 0.6601                & \cellcolor[HTML]{FFCCC9}{{0.7069}}  & 0.6010                            & \cellcolor[HTML]{C9E4FF}{{0.6724}}  \\
                                  &                                 & TNR  & 0.6053          & 0.6184           & 0.6579        & 0.6579                            & 0.6579                & \cellcolor[HTML]{FFCCC9}{{0.7105}}  & 0.6184                            & \cellcolor[HTML]{C9E4FF}{{0.6711}}  \\
                                  &                                 & ACC  & 0.5954          & 0.6120           & 0.6577        & 0.6432                            & 0.6598                & \cellcolor[HTML]{FFCCC9}{{0.7075}}  & 0.6037                            &\cellcolor[HTML]{C9E4FF}{{0.6722}}  \\ 
\cline{2-11}
                                  & \multirow{4}{*}{\textbf{Average}} & AUC & 0.6648          & 0.6816           & 0.7403        & \cellcolor[HTML]{C9E4FF}{{0.7656}} & 0.7264                & 0.7550                             & 0.7409                            & \cellcolor[HTML]{FFCCC9}{{0.7738}}   \\
                                  &                                 & TPR  & 0.6034          & 0.6207           & 0.6690        & \cellcolor[HTML]{C9E4FF}{{0.6900}} & 0.6647                & 0.6766                            & 0.6678                            & \cellcolor[HTML]{FFCCC9}{{0.7016}}   \\
                                  &                                 & TNR   & 0.6109          & 0.6247           & 0.6673        & \cellcolor[HTML]{C9E4FF}{{0.7004}} & 0.6683                & 0.6839                            & 0.6722                            & \cellcolor[HTML]{FFCCC9}{{0.7004}}   \\
                                  &                                 & ACC   & 0.6055          & 0.6212           & 0.6675        & \cellcolor[HTML]{C9E4FF}{{0.6940}} & 0.6669                & 0.6794                            & 0.6674                            & \cellcolor[HTML]{FFCCC9}{{0.7010}}   \\ 
\cline{2-11}
                                  & \multirow{4}{*}{\textbf{Overall}} & AUC & 0.8005          & 0.8062           & 0.8264        & \cellcolor[HTML]{C9E4FF}{{0.8445}} & 0.8258                & 0.8394                            & 0.8225                            & \cellcolor[HTML]{FFCCC9}{{0.8491}}   \\
                                  &                                 & TPR   & 0.7300          & 0.7285           & 0.7374        & \cellcolor[HTML]{C9E4FF}{{0.7567}}                            & 0.7270                & 0.7329                            & 0.7270 & \cellcolor[HTML]{FFCCC9}{{0.7596}}   \\
                                  &                                 & TNR   & 0.7311          & 0.7288           & 0.7382        & \cellcolor[HTML]{C9E4FF}{{0.7618}}                            & 0.7288                & 0.7358                            & 0.7288 & \cellcolor[HTML]{FFCCC9}{{0.7618}}   \\
                                  &                                 & ACC   & 0.7304          & 0.7286           & 0.7377        & \cellcolor[HTML]{C9E4FF}{{0.7587}}                            & 0.7277                & 0.7341                            & 0.7277 & \cellcolor[HTML]{FFCCC9}{{0.7605}}   \\ 
\hline
\hline
\multirow{16}{*}{\rotatebox[origin=c]{90}{\textbf{Unseen Domain}}} & \multirow{4}{*}{\textbf{TOMMY2}}   & AUC & 0.6298          & 0.6466           & 0.6760        & 0.6798                            & 0.6714                & 0.6919                            & \cellcolor[HTML]{C9E4FF}{{0.6994}} & \cellcolor[HTML]{FFCCC9}{{0.7288}}   \\
                                  &                                 & TPR   & 0.5954          & 0.6029           & 0.6248        & 0.6314                            & 0.6157                & 0.6271                            & \cellcolor[HTML]{C9E4FF}{{0.6461}} & \cellcolor[HTML]{FFCCC9}{{0.6769}}   \\
                                  &                                 & TNR   & 0.5939          & 0.6041           & 0.6269        & 0.6345                            & 0.6168                & 0.6294                            & \cellcolor[HTML]{C9E4FF}{{0.6447}} & \cellcolor[HTML]{FFCCC9}{{0.6777}}   \\
                                  &                                 & ACC   & 0.5951          & 0.6031           & 0.6251        & 0.6319                            & 0.6159                & 0.6275                            & \cellcolor[HTML]{C9E4FF}{{0.6459}} & \cellcolor[HTML]{FFCCC9}{{0.6771}}   \\ 
\cline{2-11}
                                  & \multirow{4}{*}{\textbf{INBreast}} & AUC & 0.4692          & 0.5195           & 0.6522        & 0.6791                            & 0.7097                & \cellcolor[HTML]{C9E4FF}{{0.7649}} & 0.6623                            & \cellcolor[HTML]{FFCCC9}{{0.7889}}   \\
                                  &                                 & TPR   & 0.4080          & 0.5200           & 0.5520        & \cellcolor[HTML]{C9E4FF}{{0.7120}} & 0.6080                & 0.7040                            & 0.5760                            & \cellcolor[HTML]{FFCCC9}{{0.7520}}   \\
                                  &                                 & TNR   & 0.4348          & 0.5217           & 0.5870        & 0.6304                            & 0.6304                & \cellcolor[HTML]{C9E4FF}{{0.6957}} & 0.5870                            & \cellcolor[HTML]{FFCCC9}{{0.6957}}   \\
                                  &                                 &ACC   & 0.4152          & 0.5205           & 0.5614        & 0.6901                            & 0.6140                & \cellcolor[HTML]{C9E4FF}{{0.6998}} & 0.5789                            & \cellcolor[HTML]{FFCCC9}{{0.7368}}   \\ 
\cline{2-11}
                                  & \multirow{4}{*}{\textbf{Average}} & AUC & 0.5495          & 0.5831           & 0.6641        & 0.6795                            & 0.6906                & \cellcolor[HTML]{C9E4FF}{{0.7284}} & 0.6809                            & \cellcolor[HTML]{FFCCC9}{{0.7589}}   \\
                                  &                                 & TPR   & 0.5017          & 0.5615           & 0.5884        & \cellcolor[HTML]{C9E4FF}{{0.6717}} & 0.6119                & 0.6656                            & 0.6111                            & \cellcolor[HTML]{FFCCC9}{{0.7145}}   \\
                                  &                                 & TNR   & 0.5144          & 0.5629           & 0.6070        & 0.6325                            & 0.6236                & \cellcolor[HTML]{C9E4FF}{{0.6626}} & 0.6159                            & \cellcolor[HTML]{FFCCC9}{{0.6867}}   \\
                                  &                                 & ACC   & 0.5052          & 0.5618           & 0.5933        & 0.661                             & 0.6150                & \cellcolor[HTML]{C9E4FF}{{0.6637}} & 0.6124                            & \cellcolor[HTML]{FFCCC9}{{0.7070}}   \\ 
\cline{2-11}
                                  & \multirow{4}{*}{\textbf{Overall}} & AUC & 0.6343          & 0.6494           & 0.6784        & 0.6792                            & 0.6750                & \cellcolor[HTML]{C9E4FF}{{0.6955}} & 0.6979                            & \cellcolor[HTML]{FFCCC9}{{0.7341}}   \\
                                  &                                 & TPR   & 0.5979          & 0.6082           & 0.6229        & \cellcolor[HTML]{C9E4FF}{{0.6341}} & 0.6238                & 0.6301                            & 0.6413                            & \cellcolor[HTML]{FFCCC9}{{0.6767}}   \\
                                  &                                 & TNR   & 0.6000          & 0.6091           & 0.6250        & \cellcolor[HTML]{C9E4FF}{{0.6341}} & 0.6250                & 0.6295                            & 0.6432                            & \cellcolor[HTML]{FFCCC9}{{0.6773}}   \\
                                  &                                 & ACC   & 0.5982          & 0.6083           & 0.6233        & \cellcolor[HTML]{C9E4FF}{{0.6341}} & 0.6240                & 0.6300                            & 0.6416                            & \cellcolor[HTML]{FFCCC9}{{0.6768}}   \\ 
\hline
\hline
\multicolumn{2}{c|}{\multirow{4}{*}{\textbf{Average}}}                 & AUC & 0.6187          & 0.6422           & 0.7098        & 0.7312                            & 0.7120                & \cellcolor[HTML]{C9E4FF}{{0.7444}} & 0.7169                            & \cellcolor[HTML]{FFCCC9}{{0.7678}}   \\
\multicolumn{2}{c|}{}                                                 & TPR & 0.5627          & 0.5970           & 0.6367        & \cellcolor[HTML]{C9E4FF}{{0.6827}} & 0.6435                & 0.6722                            & 0.6451                            & \cellcolor[HTML]{FFCCC9}{{0.7067}}   \\
\multicolumn{2}{c|}{}                                                &TNR  & 0.5723          & 0.6000           & 0.6431        & 0.6732                            & 0.6504                & \cellcolor[HTML]{C9E4FF}{{0.6754}} & 0.6497                            & \cellcolor[HTML]{FFCCC9}{{0.6949}}   \\
\multicolumn{2}{c|}{}                                                 &ACC & 0.5654          & 0.5975           & 0.6378        & \cellcolor[HTML]{C9E4FF}{{0.6808}} & 0.6461                & 0.6731                            & 0.6454                            & \cellcolor[HTML]{FFCCC9}{{0.7034}}   \\ 
\hline
\multicolumn{2}{c|}{\multirow{4}{*}{\textbf{Overall}}}                &AUC & 0.7386          & 0.7476           & 0.7646        & 0.7702                            & 0.7634                & \cellcolor[HTML]{C9E4FF}{{0.7806}} & 0.7654                            & \cellcolor[HTML]{FFCCC9}{{0.8013}}   \\
\multicolumn{2}{c|}{}                                                &TPR  & 0.6735          & 0.6735           & 0.6859        & \cellcolor[HTML]{C9E4FF}{{0.7049}} & 0.6935                & 0.6959                            & 0.6945                            & \cellcolor[HTML]{FFCCC9}{{0.7258}}   \\
\multicolumn{2}{c|}{}                                                &TNR  & 0.6736          & 0.6736           & 0.6863        & \cellcolor[HTML]{C9E4FF}{{0.7049}} & 0.6956                & 0.6968                            & 0.6956                            & \cellcolor[HTML]{FFCCC9}{{0.7269}}   \\
\multicolumn{2}{c|}{}                                                &ACC  & 0.6736          & 0.6736           & 0.686         & \cellcolor[HTML]{C9E4FF}{{0.7049}} & 0.6940                & 0.6961                            & 0.6948                            & \cellcolor[HTML]{FFCCC9}{{0.7261}}   \\
\hline
\hline
\end{tabular}
    
    }
  \label{tab:main}
\end{table*}

 \paragraph{Vendor-Specific Mammography Scanner Information.} In our study, we utilised four distinct mammography datasets collected from different scanners to examine the impact of scanner variability on mammography analysis. 
The CBIS-DDSM dataset was digitalized with four different scanners: DBA scanner at MGH, HOWTEK scanner at MGH, LUMISYS scanner at Wake Forest University, and HOWTEK scanner at ISMD. Additional information about this dataset can be found here \footnote{\url{http://www.eng.usf.edu/cvprg/mammography/database.html}}. 
The CMMD dataset was acquired on a GE Senographe DS mammography system.
The InBreast dataset was captured with MammoNovation Siemens FFDM equipment at the Breast Centre in CHSJ, Porto \citep{moreira2012inbreast}.
Lastly, the TOMMY dataset was collected by a commercially available (Hologic) digital mammography system \citep{gilbert2011tommy}.
By analysing and processing these diverse datasets, we aim to investigate the generalisation ability of the proposed MammoDG framework under the influence of scanner characteristics on mammography and explore potential implications for clinical applications.

\subsection{Performance Evaluation}
To quantitatively evaluate the performance of our method, we adopt four popular classification metrics for all experiments, \ie, the area under receiver operator characteristic curve (AUC), true positive recall (TPR), true negative recall (TNR) and accuracy (ACC). All models are trained on the training sets of seen domains and evaluated on the test set of each domain, respectively. It is worth noting that, to obtain average performance, we simply average the metric values of all the target domains, \ie, different thresholds are adopted across domains. For overall performance, we aggregate the test set of all target domains and then evaluate the model on the mixed test set, \ie, the same threshold is adopted for the classification of all domains.

\begin{figure}[t!]%
\centering
\includegraphics[width=0.9\textwidth]{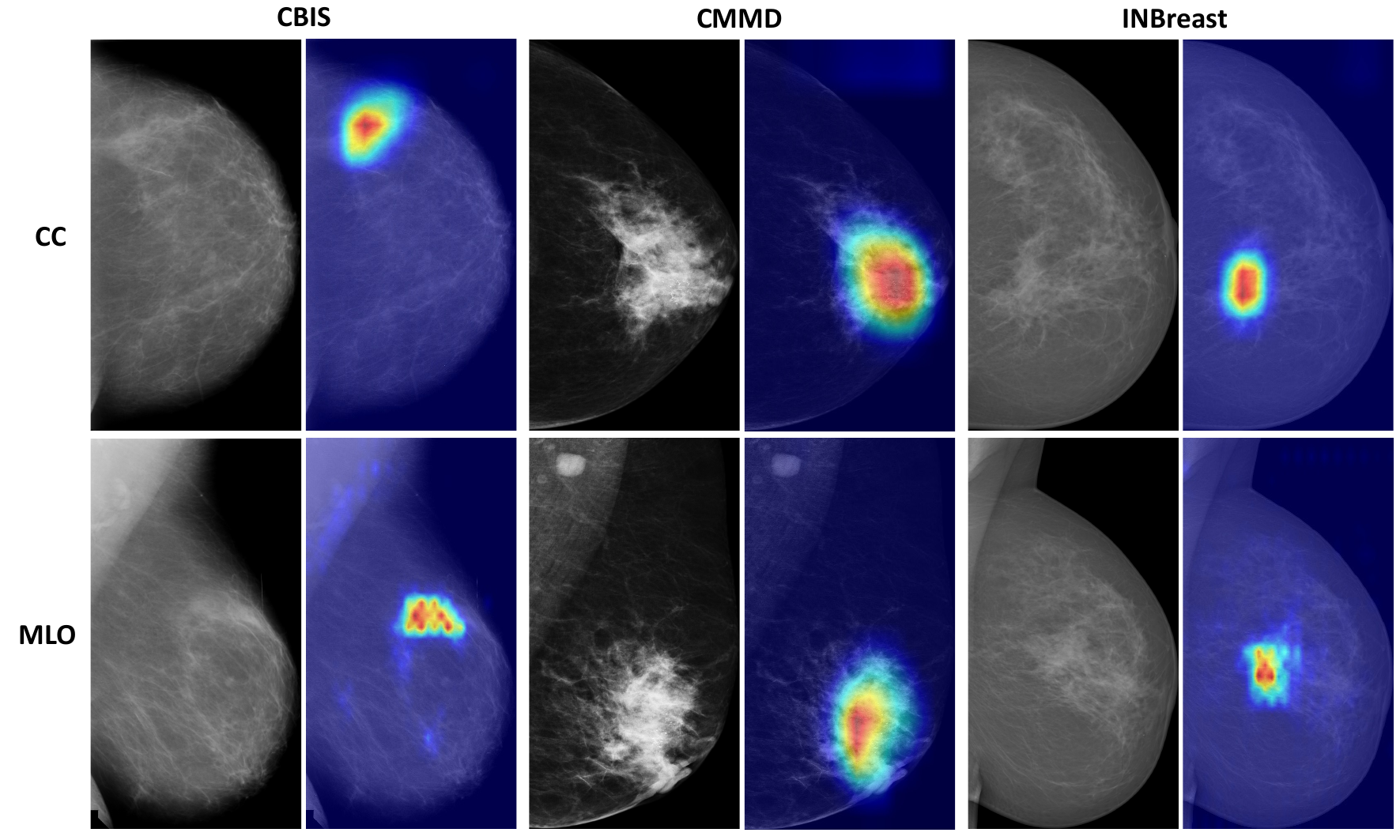}
\caption{The visualisation of heatmap after cross-channel cross-view enhancement. Three malignant cases, one from each dataset, are tested. While the red and blue regions denote abnormal and normal tissues with high confidence. It is important to note that the yellow regions represent abnormal tissues with low confidence, and therefore should be verified through additional scrutiny.
}
\label{fig:cve_attentionmap}
\end{figure}

\subsection{Comparison with state-of-the-art methods}

We compare our network against state-of-the-art mammography classification methods, including BIRADS~\citep{geras2017high}, DMV-CNN~\citep{wu2019deep}, MVFF~\citep{khan2019multi} and GMIC~\citep{shen2021interpretable}. To further demonstrate the generalisation ability of our model, we also reimplement a generalizable mammography detection framework MSVCL~\citep{li2021domain} in two ways. MSVCL(ResNet), MSVCL(FCOS) both utilize ResNet as the backbone but the latter additionally incorporates feature pyramid network to leverage multi-level features as FCOS~\citep{tian2019fcos} does.
BIRADS, DMV-CNN and MVFF are designed in the multi-view fashion while GMIC and MSVCL are the single-view frameworks. For a fair comparison, we obtain breast-level predictions by averaging their image-level predictions for those single-view frameworks.
As displayed in Table~\ref{tab:main}, our framework produces superior performance on both Seen domain and Unseen domain. For Seen domain, our MammoDG surpasses the second-best method GMIC by 0.0082 in AUC and 0.0070 in ACC of the average performance, 0.0045 in AUC and 0.0018 in ACC of the overall performance, respectively. For Unseen domain, our method improves the average and overall performance by a considerable margin of 0.0305 in AUC and 0.0433 in ACC, 0.0386 in AUC and 0.0468 in ACC, respectively, than the generalisable method MSVCL(FCOS). The consistent improvements on all datasets result in superb advances in all domains in view of all four metrics of the average and overall performance.

\subsection{Ablation Studies}
\begin{table*}[!tbp]
  \centering
  \caption{Quantitatively ablation studies on our domain generalisation setting. The public datasets CBIS and CMMD are treated as Seen domain while INBreast is treated as Unseen domain. The module ``\textbf{CVE}'' means Cross-Channel Cross-View Enhancement while the module ``\textbf{MS}'' denotes Mixstyle. ``\textbf{GE}'' denotes Global Encoder while ``\textbf{MICL}'' represents Multi-Instance Contrastive Learning. The top values are bold.}
    \resizebox{0.8\textwidth}{!}{%

\begin{tabular}{c|c|c|ccccc} 
\hline
\hline
\multicolumn{2}{c|}{\textbf{Method }}   &      \textbf{Metrics}                         & \textbf{Baseline} & \textbf{+CVE} & \textbf{+MS}    & \textbf{+GE}    &  \textbf{+MICL}    \\ 
\hline
\multirow{16}{*}{\textbf{Seen }}  & \multirow{4}{*}{\textbf{CBIS}}  & AUC   & 0.6928            & 0.7066        & 0.7590          & 0.6960                 & \textbf{0.7602}  \\
                                  &            &TPR                        & 0.6144            & 0.6275        & 0.6863          & 0.6340          & \textbf{0.6928}           \\
                                  &          &TNR                          & 0.6275            & 0.6471        & \textbf{0.6961} & 0.6569                  & 0.6863           \\
                                  &          &ACC                          & 0.6196            & 0.6353        & 0.6902          & 0.6431          & \textbf{0.6902}           \\ 
\cline{2-8}
                                  & \multirow{4}{*}{\textbf{CMMD}} &AUC    & 0.7881            & 0.7926        & 0.7840          & \textbf{0.8567}         & 0.8309           \\
                                  &          &TPR                          & 0.7043            & 0.7217        & 0.7043          & \textbf{0.7652}          & 0.7391           \\
                                  &          &TNR                          & 0.7033            & 0.7317        & 0.7114          & \textbf{0.7805}          & 0.7440           \\
                                  &          &ACC                          & 0.7036            & 0.7285        & 0.7092          & \textbf{0.7756}         & 0.7424           \\ 
\cline{2-8}
                                  & \multirow{4}{*}{\textbf{Average }} &AUC & 0.7405            & 0.7496        & 0.7715          & 0.7764                 & \textbf{0.7956}  \\
                                  &            &TPR                        & 0.6594            & 0.6746        & 0.6953          & 0.6996        & \textbf{0.7160}           \\
                                  &           &TNR                         & 0.6654            & 0.6894        & 0.7038          & \textbf{0.7187}        & 0.7152           \\
                                  &           &ACC                         & 0.6616            & 0.6819        & 0.6997          & 0.7094         & \textbf{0.7163}           \\ 
\cline{2-8}
                                  & \multirow{4}{*}{\textbf{Overall }} &AUC& 0.7777            & 0.7869        & 0.7983          & 0.8037                   & \textbf{0.8201}  \\
                                  &           &TPR                         & 0.7052            & 0.7239        & 0.7089          & 0.7164                  & \textbf{0.7463}  \\
                                  &           &TNR                         & 0.7069            & 0.7213        & 0.7126          & 0.7184                   & \textbf{0.7500}  \\
                                  &            &ACC                        & 0.7062            & 0.7224        & 0.7110          & 0.7175                  & \textbf{0.7484}  \\ 
\hline
\hline
\multirow{4}{*}{\textbf{Unseen }} & \multirow{4}{*}{\textbf{INBreast}} &AUC& 0.6048            & 0.7780        & 0.8193          & 0.8064                 & \textbf{0.8289}  \\
                                  &        &TPR                            & 0.5600            & 0.6960        & 0.7280          & 0.7200                   & \textbf{0.7920}  \\
                                  &         &TNR                           & 0.5870            & 0.7391        & 0.7391          & 0.7391                   & \textbf{0.8043}  \\
                                  &          &ACC                          & 0.5673            & 0.7076        & 0.7310          & 0.7251                 & \textbf{0.7953}  \\ 
\hline
\hline
\multicolumn{2}{c|}{\multirow{4}{*}{\textbf{Average }}}   &AUC             & 0.6952            & 0.7591        & 0.7874          & 0.7864                  & \textbf{0.8067}  \\
\multicolumn{2}{c|}{}                        &TPR                          & 0.6262            & 0.6817        & 0.7062          & 0.7064                  & \textbf{0.7413}  \\
\multicolumn{2}{c|}{}                      &TNR                            & 0.6393            & 0.7060        & 0.7155          & 0.7255                  & \textbf{0.7449}  \\
\multicolumn{2}{c|}{}                     &ACC                             & 0.6302            & 0.6905        & 0.7101          & 0.7146                   & \textbf{0.7426}  \\ 
\hline
\multicolumn{2}{c|}{\multirow{4}{*}{\textbf{Overall }}}     &AUC           & 0.7781            & 0.7993        & 0.8207          & 0.8213                  & \textbf{0.8364}  \\
\multicolumn{2}{c|}{}                    &TPR                              & 0.6997            & 0.7201        & 0.7354          & 0.7455                 & \textbf{0.7659}  \\
\multicolumn{2}{c|}{}                     &TNR                             & 0.7081            & 0.7234        & 0.7386          & 0.7487                   & \textbf{0.7691}  \\
\multicolumn{2}{c|}{}                    &ACC                              & 0.7039            & 0.7217        & 0.7370          & 0.7471                   & \textbf{0.7675}  \\
\hline
\hline
\end{tabular}

    }
  \label{tab:ablation_main}
\end{table*}

In this part, we conduct extensive ablation studies on the publicly available datasets, \ie, CBIS, CMMD as Seen domain, and INBreast as Unseen domain.
\paragraph{The Effectiveness of Each Module. }
As shown in Table~\ref{tab:ablation_main}, we validate the effectiveness of each module in our framework. The Baseline model consists of two Resnet18 branches where CC and MLO views are encoded respectively and then fused by concatenating two feature embeddings for the final breast-level prediction. The Cross-Channel Cross-View Enhancement module (+CVE) dramatically advances the overall performance over the Baseline model by 0.0212 in AUC and 0.0178 in ACC. The Mixstyle augmentation strategy (+MS) is further incorporated at each stage to mitigate the domain shift problem and achieve significant improvement, particularly on unseen domains. While the Global Encoder (+GE) explores the shared representation of two views, the Multi-Instance Contrastive Learning strategy (+MICL) conducts view-specific learning and endows the full MammoDG model with the overall performance of 0.8364, 0.7659, 0.7691, 0.7675 in AUC, TPR, TNR, ACC.

\paragraph{Details of CVE Module. }
We quantitatively explore the efficacy of each component in our CVE module in Table~\ref{tab:cve_part} and conduct the experiments based on the full MammoDG. The model without the entire CVE module achieves an overall performance of 0.8014, 0.7141 in AUC, ACC, respectively. Cross-Channel Enhancement (CE) brings 0.0300 AUC gains and 0.0280 ACC gains in overall performance while Cross-View Enhancement (VE) further improves by 0.0050 and 0.0254, respectively. To qualitatively verify the effectiveness of our CVE module, we visualize the heatmap of three samples after manipulating cross-channel cross-view enhancement. Figure~\ref{fig:cve_attentionmap} clearly demonstrates that our method successfully detects malignant tissues after these enhancements.

\paragraph{Discussion on view-specific learning. }
In Table~\ref{tab:micl_part}, we conduct experiments on different strategies of view-specific learning. We first replace our MICL with a vanilla classifier head to give supervision on image-level classification, which degrades the overall performance by 0.0128 and 0.0227 in AUC and ACC. Additionally, we replace our MICL with a MIL aggregator~\cite{li2021dual}, leading to a significant drop of 0.0117 and 0.0166 in overall AUC and ACC.

\paragraph{Discussion on the balancing hyper-parameters. }
We discuss the best choice of the balancing ratio of breast-level prediction and image-level predictions in Table~\ref{tab:ensemble}. The equal weight (1:1:1) for CC, MLO and breast-level predictions achieves the best performance. We also discuss the balancing hyper-parameter $\lambda$ to weight supervision loss and contrastive loss in MICL in Table~\ref{tab:micl_lam}. $\lambda$ should be set as 0.5 to obtain the best overall AUC and ACC.

\paragraph{Details of the MICL strategy. }
In Table~\ref{tab:micl_nn}, we explore the effects of the number of tiles (instances) in one bag on our MICL strategy. The experiment results show that when one image is divided into $4\times4$ tiles, the best overall performance is achieved.


\begin{table*}[!t]
\small
\begin{subtable}{0.5\linewidth}
\centering
\captionsetup{width=0.9\linewidth}
        \begin{tabular}{c|c|c|c} 
\hline
\hline
\textbf{}                                        & \textbf{w/o CVE} & \textbf{CE}     & \textbf{CVE} (ours)     \\ 
\hline
\multirow{4}{*}{\textbf{Seen}}                       & 0.8070           & 0.8035          & \textbf{0.8201}  \\
                                                       & 0.7276           & 0.7164          & \textbf{0.7463}  \\
                                                       & 0.7328           & 0.7241          & \textbf{0.7500}  \\
                                                       & 0.7305           & 0.7208          & \textbf{0.7484}  \\ 
\hline
\multirow{4}{*}{\textbf{Unseen}}                         & 0.7012           & \textbf{0.8355} & 0.8289           \\
                                                       & 0.6080           & 0.7360          & \textbf{0.7920}  \\
                                                       & 0.6304           & 0.7826          & \textbf{0.8043}  \\
                                                       & 0.6140           & 0.7485          & \textbf{0.7953}  \\ 
\hline
\hline
\multicolumn{1}{l|}{\multirow{4}{*}{\textbf{Overall}}} & 0.8014           & 0.8314          & \textbf{0.8364}  \\
\multicolumn{1}{l|}{}                                  & 0.7125           & 0.7430          & \textbf{0.7659}  \\
\multicolumn{1}{l|}{}                                  & 0.7157           & 0.7411          & \textbf{0.7691}  \\
\multicolumn{1}{l|}{}                                  & 0.7141           & 0.7421          & \textbf{0.7675}  \\
\hline
\hline
\end{tabular}
        \caption{\textbf{Each component in CVE module.} CE brings 0.0300 AUC gains and 0.0280 ACC gains in overall performance while VE further improves by 0.0050 and 0.0254, respectively.}
        \label{tab:cve_part}
    \end{subtable}
\begin{subtable}{0.5\linewidth}
\centering
\captionsetup{width=0.9\linewidth}
\begin{tabular}{c|c|c|c} 
\hline
\hline
\textbf{}                                        & \textbf{w/o ens} & \textbf{1:1:2}  & \textbf{1:1:1} (ours)   \\ 
\hline
\multirow{4}{*}{\textbf{Seen}}                       & 0.8230           & \textbf{0.8245} & 0.8201           \\
                                                       & 0.7388           & 0.7425          & \textbf{0.7463}  \\
                                                       & 0.7471           & 0.7443          & \textbf{0.7500}  \\
                                                       & 0.7435           & 0.7435          & \textbf{0.7484}  \\ 
\hline
\multirow{4}{*}{\textbf{Unseen}}                         & 0.8219           & 0.8221          & \textbf{0.8289}  \\
                                                       & 0.7520           & 0.7520          & \textbf{0.7920}  \\
                                                       & 0.8043           & 0.8043          & \textbf{0.8043}  \\
                                                       & 0.7661           & 0.7661          & \textbf{0.7953}  \\ 
\hline
\hline
\multicolumn{1}{l|}{\multirow{4}{*}{\textbf{Overall}}} & 0.8306           & 0.8317          & \textbf{0.8364}  \\
\multicolumn{1}{l|}{}                                  & 0.7532           & 0.7608          & \textbf{0.7659}  \\
\multicolumn{1}{l|}{}                                  & 0.7563           & 0.7614          & \textbf{0.7691}  \\
\multicolumn{1}{l|}{}                                  & 0.7548           & 0.7612          & \textbf{0.7675}  \\
\hline
\hline
\end{tabular}
        \caption{\textbf{The balancing ratio of breast-level and image-level predictions. } The equal weight for CC, MLO and breast-level predictions promise the best performance.}
        \label{tab:ensemble}
    \end{subtable}
\begin{subtable}{0.5\linewidth}
\centering
\captionsetup{width=0.9\linewidth}
\begin{tabular}{c|c|c|c} 
\hline
\hline
\textbf{}                                        & \textbf{vanilla} & \textbf{MIL}    & \textbf{MICL} (ours)            \\ 
\hline
\multirow{4}{*}{\textbf{Seen}}                       & 0.8040                      & 0.8098 & \textbf{0.8201}  \\
                                                       & 0.7276                      & 0.7351 & \textbf{0.7463}  \\
                                                       & 0.7328                      & 0.7385 & \textbf{0.7500}  \\
                                                       & 0.7305                      & 0.7370 & \textbf{0.7484}  \\ 
\hline
\multirow{4}{*}{\textbf{Unseen}}                         & 0.8217                      & 0.8266 & \textbf{0.8289}  \\
                                                       & 0.7440                      & 0.7600 & \textbf{0.7920}  \\
                                                       & 0.7609                      & 0.7826 & \textbf{0.8043}  \\
                                                       & 0.7485                      & 0.7661 & \textbf{0.7953}  \\ 
\hline
\hline
\multicolumn{1}{l|}{\multirow{4}{*}{\textbf{Overall}}} & 0.8236                      & 0.8247 & \textbf{0.8364}  \\
\multicolumn{1}{l|}{}                                  & 0.7431                      & 0.7481 & \textbf{0.7659}  \\
\multicolumn{1}{l|}{}                                  & 0.7463                      & 0.7538 & \textbf{0.7691}  \\
\multicolumn{1}{l|}{}                                  & 0.7448                      & 0.7509 & \textbf{0.7675}  \\
\hline
\hline
\end{tabular}
        \caption{\textbf{Vanilla classifier vs. MICL as view-specific decoders. }MICL improves the AUC and ACC in overall performance by 0.0128 and 0.0227.}
        \label{tab:micl_part}
    \end{subtable}
\begin{subtable}{0.5\linewidth}
\centering
\captionsetup{width=0.9\linewidth}
\begin{tabular}{c|c|c|c} 
\hline
\hline
$\lambda$                                      & \textbf{0.2}    & \textbf{0.5} (ours)   & \textbf{1.0}  \\ 
\hline
\multirow{4}{*}{\textbf{Seen}}                       & 0.8137          & \textbf{0.8201} & 0.8146        \\
                                                       & 0.7201          & \textbf{0.7463} & 0.7276        \\
                                                       & 0.7270          & \textbf{0.7500} & 0.7299        \\
                                                       & 0.7240          & \textbf{0.7484} & 0.7289        \\ 
\hline
\multirow{4}{*}{\textbf{Unseen}}                         & \textbf{0.8369} & 0.8289          & 0.7821        \\
                                                       & 0.7760          & \textbf{0.7920} & 0.5840        \\
                                                       & 0.7826          & \textbf{0.8043} & 0.7174        \\
                                                       & 0.7778          & \textbf{0.7953} & 0.6199        \\ 
\hline
\hline
\multicolumn{1}{l|}{\multirow{4}{*}{\textbf{Overall}}} & 0.8345          & \textbf{0.8364} & 0.8313        \\
\multicolumn{1}{l|}{}                                  & \textbf{0.7710} & 0.7659          & 0.7481        \\
\multicolumn{1}{l|}{}                                  & 0.7640          & \textbf{0.7691} & 0.7538        \\
\multicolumn{1}{l|}{}                                  & 0.7675          & \textbf{0.7675} & 0.7509        \\
\hline
\hline
\end{tabular}
        \caption{\textbf{The balancing hyper-parameter $\lambda$. } The best value to weight cross entropy loss and contrastive loss in MICL is 0.5 with best AUC and ACC.}
        \label{tab:micl_lam}
    \end{subtable}
\begin{subtable}{\linewidth}
\centering
\captionsetup{width=\linewidth}
\begin{tabular}{c|c|c|c|c} 
\hline
\hline
\textbf{$n$}                                             & \textbf{3} & \textbf{4} (ours)      & \textbf{5}      & \textbf{6}  \\ 
\hline
\multirow{4}{*}{\textbf{Seen}}                       & 0.8178     & 0.8201          & \textbf{0.8254} & 0.8093      \\
                                                       & 0.7463     & 0.7463          & \textbf{0.7500} & 0.7276      \\
                                                       & 0.7443     & 0.7500          & \textbf{0.7500} & 0.7328      \\
                                                       & 0.7453     & 0.7484          & \textbf{0.7500} & 0.7305      \\ 
\hline
\multirow{4}{*}{\textbf{Unseen}}                         & 0.8241     & \textbf{0.8289} & 0.8231          & 0.8270      \\
                                                       & 0.7520     & 0.7920          & \textbf{0.8000} & 0.7600      \\
                                                       & 0.7826     & \textbf{0.8043} & 0.7826          & 0.7826      \\
                                                       & 0.7673     & \textbf{0.7953} & 0.7953          & 0.7661      \\ 
\hline
\hline
\multicolumn{1}{l|}{\multirow{4}{*}{\textbf{Overall}}} & 0.8310     & \textbf{0.8364} & 0.8325          & 0.8348      \\
\multicolumn{1}{l|}{}                                  & 0.7543     & \textbf{0.7659} & 0.7583          & 0.7659      \\
\multicolumn{1}{l|}{}                                  & 0.7614     & \textbf{0.7691} & 0.7614          & 0.7691      \\
\multicolumn{1}{l|}{}                                  & 0.7579     & \textbf{0.7675} & 0.7598          & 0.7675      \\
\hline
\hline
\end{tabular}
        \caption{\textbf{The number of tiles $n$ in MICL. } The overall performance achieves the best when $n=4$.}
        \label{tab:micl_nn}
    \end{subtable}
\caption{Quantitatively ablation studies on details of each parts in our MammoDG. All models are trained on the train set of CBIS and CMMD and test on INBreast and the test set of CBIS and CMMD. (Four metrics from up to bottom are AUC, TPR, TNR, ACC.)}\label{tab:1}
\end{table*}

\section{Discussion}\label{sec4}

\paragraph{MammoDG outperforms traditional supervised methods on mammography diagnosis.}
In our comparison with traditional supervised methods (''Seen" Category in Table~\ref{tab:main}) for mammography diagnosis, MammoDG demonstrated superior performance across all metrics. This is largely attributed to the effective use of multi-view mammograms framework and the innovative contrastive mechanism that enhances generalisation capabilities. Traditional models often struggle with the high variability and complex patterns found in mammograms, while MammoDG was designed to robustly manage this inherent complexity. In terms of AUC, TPR, TNR, and ACC, our method consistently outperformed traditional supervised methods, highlighting the benefit of leveraging advanced domain generalisation mechanisms for this task.

\paragraph{MammoDG consistently surpasses the generalisable mammography diagnosis methods on unseen domains.}
Another distinguishing feature of MammoDG is its ability to maintain superior performance even when tested on unseen domains. This was a limitation observed in previous studies with other generalisable mammography diagnosis methods. According to the ''Unseen" part of Table~\ref{tab:main} \&~\ref{tab:ablation_main}, MammoDG’s robustness to out-of-distribution data, collected from various vending machines and protocols, allows it to handle the data distribution shift in large cohorts effectively. This shows the feasibility of the deployment of MammoDG in real-world scenarios across various centres and hospitals.

\paragraph{MammoDG saves the cost of annotation in target domains.}
MammoDG’s ability to achieve high performance with limited annotations is crucial to the medical image analysis community. Given the difficulty and expense of acquiring reliable annotations, a model that can still excel with such limitations is invaluable. As compared to traditional supervised models that require extensive and costly annotations for training, MammoDG substantially cuts the cost of annotation in target domains, which makes it an efficient and cost-effective solution for large-scale mammography analysis across multiple centers.

\paragraph{MammoDG provides reliable evidence for clinical decisions.}
%
The results from this study have significant practical implications for the healthcare industry, specifically for radiologists and healthcare providers engaged in breast cancer detection. The machine learning model developed in this research demonstrated robust performance across various datasets, with promising implications for real-world application.
In the domain of breast cancer diagnosis, MammoDG is an especially powerful tool as it considers both CC and MLO views, providing a comprehensive analysis that leverages cross-view complementary information. As depicted in Figure~\ref{fig:cve_attentionmap}, MammoDG consistently generates reliable attention regions, providing evidence that matches well with radiologists’ diagnoses. The intersection over union between our model's attention regions and the areas highlighted by radiologists consistently exceeded a threshold, indicating MammoDG's capability to provide trustworthy and actionable insights for clinical decisions.

In the future, we aim to conduct reader studies to measure the extent to which accuracy improves when radiologists use our system and to evaluate their level of trust in it. Given the potential benefits of AI assistance, particularly for less-experienced readers, further investigation will be valuable in comparing the benefits of this system for both sub-specialists and community radiologists who might be called on to do this work only occasionally.

\paragraph{MammoDG's Limitations}
Despite its strengths, this study also has several limitations. First, although the model was evaluated on several diverse datasets, these are primarily from China and the UK. Additional validation on datasets from other regions and ethnicities would be valuable in assessing the global applicability of our model. Second, the results of this study are contingent on the accuracy of the ground truth labels, which are based on human interpretation and thus subject to inter-observer variability. Lastly, while the model demonstrated strong performance in distinguishing between benign and malignant cases, there remains a need to further investigate its efficacy in detecting early stage cancers, as this is crucial for improving patient outcomes. Future work should aim to address these limitations, refine the model's capabilities, and assess its performance in a real-world clinical setting.

\section{Conclusion}\label{sec:conclusion}

Our work exhibits our ability to develop a pioneering deep-learning framework for generalisable, robust, and reliable analysis of cross-domain multi-center mammography data. Our framework MammoDG outperforms traditional models when trained on limited data. We are able to provide a generalisable network that performs comparably to radiologists on breast cancer analysis without requiring specific training when transferring to new clinical sites. Extensive experiments further validate the critical importance of domain generalisation for trustworthy mammography analysis in the presence of imaging protocol variations.

\section{Check List Information}

\section*{Data availability}\label{sec5}


This study involved four datasets.
Three of them are published data and the remaining one is the private dataset.
The CBIS dataset is Breast Cancer Image Dataset from Kaggle (\url{https://www.kaggle.com/datasets/awsaf49/cbis-ddsm-breast-cancer-image-dataset}), and CMMD is The Chinese Mammography Database from \url{https://wiki.cancerimagingarchive.net/pages/viewpage.action?pageId=70230508}, and InBreast dataset from Kaggle (\url{https://www.kaggle.com/datasets/martholi/inbreast}).
The TOMMY dataset \citep{gilbert2015tommy} is not currently permitted for public release by their respective Institutional Review Boards.

\section*{Code availability}\label{sec6}
The code for this project, including all libraries used and their versions, is available online at \url{https://github.com/need update}.


\section*{Acknowledgements}\label{sec8}
LL gratefully acknowledges the financial support
from a GSK scholarship and a Girton College Graduate
Research Fellowship at the University of Cambridge.
FJG acknowledges support by the NIHR Cambridge Biomedical Research Centre and an early detection programme grant from Cancer Research UK.
AIAR acknowledges support from CMIH and CCIMI,
University of Cambridge, ESPRC Digital Core Capability Award.
CBS acknowledges the Philip Leverhulme Prize, the EPSRC fellowship EP/V029428/1, EPSRC grants EP/T003553/1, EP/N014588/1, Wellcome Trust 215733/Z/19/Z and
221633/Z/20/Z, Horizon 2020 No. 777826 NoMADS and the CCIMI.

\bibliographystyle{johd}
\bibliography{bib}

\clearpage
\section*{Supplementary Material}
\bigskip 
This supplementary document provides additional information and visual results to complement the main paper, aiming to elaborate on the practical aspects and offer further insights into our approach and experimental findings.

\section{AUROC Curves, Data Distribution \& Further Statistics}
\begin{figure}[h!]%
\centering
\includegraphics[width=0.8\textwidth]{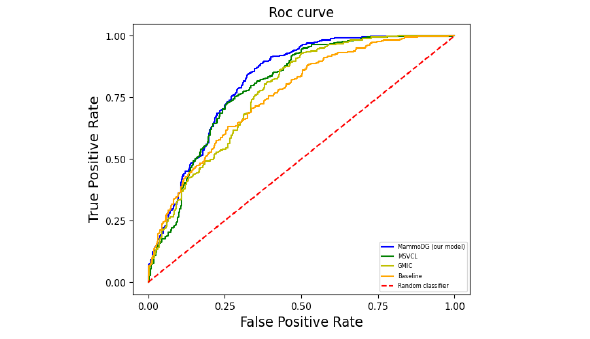}
\caption{Receiver Operating Characteristic (ROC) curves comparing our technique with existing methods on the test set of the Seen Domains and all data of the Unseen Domains.}
\label{fig:roc_curve}
\end{figure}

To strengthen the validity of our findings, we present the receiver operating characteristic (ROC) curves comparing our technique with existing methods. The results, depicted in Fig.~\ref{fig:roc_curve}, showcase a collection of ROC curves across the test set of the Seen Domains and all data of the Unseen Domains. These curves serve as a powerful visual representation that illustrates the balance between clinical sensitivity and specificity. Upon careful examination of Fig.~\ref{fig:roc_curve}, it becomes evident that our model outperforms the existing networks in terms of its discriminative capacity for cancer diagnostics. MammoDG, in particular, exhibits curves that are positioned closer to the top-left corner, indicating superior performance.

\begin{figure}[h!]%
\centering
\includegraphics[width=0.8\textwidth]{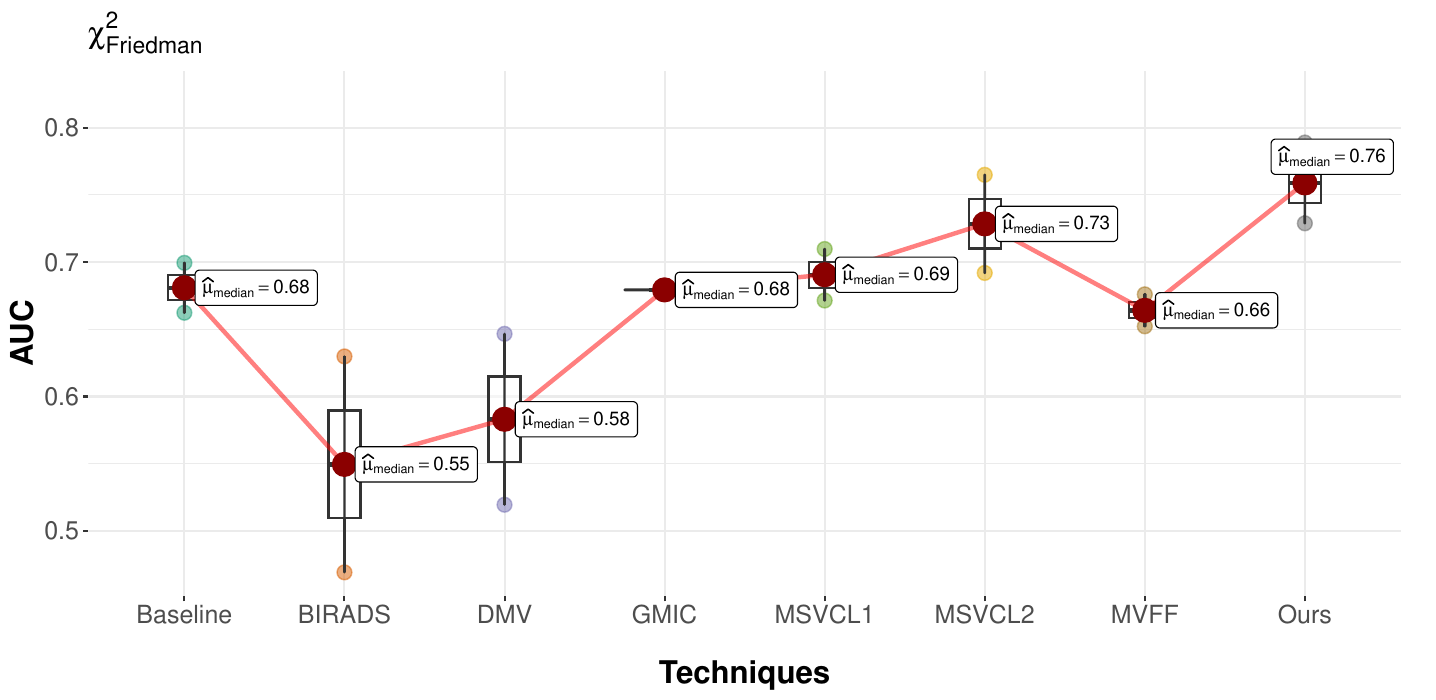}
\caption{ Statistical comparison between our technique and existing methods for Unseen Domains.}
\label{fig:stats}
\end{figure}

To strengthen the validation of our technique, we conducted a comprehensive statistical analysis displayed in Fig.~\ref{fig:stats}. Firstly, we performed a Friedman test for multiple comparisons, which allowed us to assess the overall performance of different methods. Subsequently, we employed a pair-wise comparison using the Wilcoxon test to further support our findings.

The results of these tests revealed that our technique exhibited significantly elevated performance, particularly in the case of unseen domains. This emphasises the robustness and effectiveness of our approach, showcasing its ability to generalise well beyond the training data and adapt to previously unseen domains.

\begin{figure}[h!]%
\centering
\includegraphics[width=0.8\textwidth]{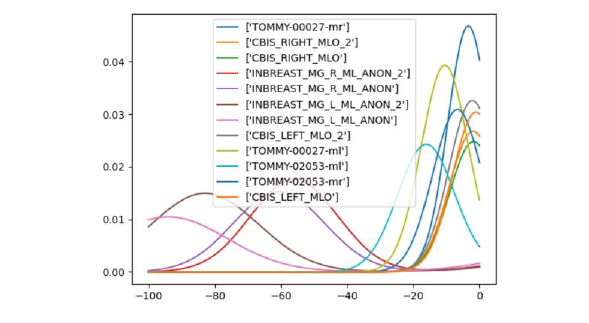}
\caption{Intensity Distribution in Datasets from Various Sites and Vendor Machines.}
\label{fig:intensity}
\end{figure}

To provide additional evidence for the effectiveness of our MammoDG model, we highlight the significant differences in intensity between datasets obtained from various sites and vendor machines. Upon closer examination in Fig.~\ref{fig:intensity}, it becomes apparent that there is a considerable domain shift between these datasets. This poses a challenge for the learning process and necessitates the use of domain generalisation techniques capable of effectively handling these variations.

\section{Comparative Analysis: Rationale for Selected Technique}
In this section, we present a comprehensive table that includes a detailed overview of existing techniques along with their respective implementation platforms. The selected techniques for comparison are marked with a tick (\checkmark) in the table, indicating the availability of their source code. It is important to note that many existing techniques do not provide open-source code, making it impossible to directly compare our approach against them.

The selection of techniques for comparison was based on the availability of their implementation details and the accessibility of their source code. It is worth mentioning that several techniques in the literature did not report sufficient information regarding their implementation, limiting our ability to include them in the comparative analysis.

We acknowledge the significance of open-source code availability for fostering reproducibility and facilitating fair comparisons. In line with this, we affirm that the source code for our proposed approach will be made available upon acceptance, ensuring transparency and enabling researchers to validate our results and conduct further investigations.
\begin{table*}[t!]
  \centering
  \caption{A survey on deep learning-based mammography analysis literature. NYUBCS: NYU breast cancer screening dataset. Compared means whether this method is compared in this paper.}
    \resizebox{1.0\textwidth}{!}{%
    \begin{tabular}{c|c|c|c|c|c}
    \toprule
     \textbf{Method} & \textbf{Datasets} & \tabincell{c}{\textbf{ROI} \\  \textbf{Annotation}} & \textbf{Compared} & \textbf{Platform} \\
    \midrule
    \tabincell{c}{Weakly supervised localization for breast \\cancer screening (GMIC)~ \citep{shen2021interpretable} }
    & \tabincell{c}{NYUBCS\\DDSM} & - &  \checkmark & PyTorch \\
    
    \hline
    
    \tabincell{c}{Domain Generalization via Multi-style and Multi-view \\Contrastive Learning (MSVCL)~\citep{li2021domain}} & \tabincell{c}{5 datasets} & - & \checkmark & PyTorch\\
        \hline
 \tabincell{c}{Multi-view Hypercomplex Neural \\Networks~\citep{li2021domain}} &\tabincell{c}{DDSM\\INbreast}& - & - & PyTorch\\
        \hline
        
    \tabincell{c}{Region-based pooling structure \\ \citep{shu2020deep} }    & \tabincell{c}{DDSM\\INbreast} & - & - & PyTorch\\
    \hline

    \tabincell{c}{External evaluation of 3 commercial\\ AI algorithms \citep{salim2020external}} & CSAW & - & - & -\\
    \hline
    
    \tabincell{c}{Robust detection with annotation-efficient\\ deep learning \citep{lotter2021robust}} & \tabincell{c}{DDSM/OMI-DB\\private dataset} & \checkmark & - & Keras\\
    \hline
    
    \tabincell{c}{Different level analysis for lesion, \\breast, and case   \citep{mckinney2020international}} & UK/US datasets & \checkmark & - & Tensorflow\\
    \hline
    \tabincell{c}{Two stages convolutional neural \\network   \citep{kim2020changes}} & \tabincell{c}{South Korea/\\UK/US} & \checkmark & - & PyTorch\\
    \hline
    \tabincell{c}{Deep Learning to Improve Breast \\Cancer Detection   \citep{shen2019deep}} & DDSM & \checkmark & \checkmark & Keras\\
        \hline
    \tabincell{c}{Deep Neural Networks Improve Radiologists’ \\
Performance (DMV-CNN)~\citep{wu2019deep}} & NYU-V1.0 & \checkmark & \checkmark & Tensorflow\\
    \hline
     \tabincell{c}{Multi-View Feature Fusion Based \\Four Views Model (MVFF)~\citep{khan2019multi}} & \tabincell{c}{DDSM\\ mini-MIAS} & - & \checkmark & Tensorflow\\
        \hline

    \tabincell{c}{Multi-view network (BIRADS)\\ \citep{geras2017high}} & \tabincell{c}{DDSM/INbreast\\others} & - & \checkmark & PyTorch\\
    \hline
    \tabincell{c}{Deep multi-instance networks with \\sparse label assignment  \citep{zhu2017deep}}  & INbreast & - & - & Keras\\
    \hline
    
    \bottomrule
    \end{tabular}
    }
  \label{tab:ablation}
\end{table*}



\end{document}